# AUTOMATIC JOINT DAMAGE QUANTIFICATION USING COMPUTER VISION AND DEEP LEARNING

Quang N.V. Tran[1*] and Jeffery R. Roesler[2]

Department of Civil and Environmental Engineering, University of Illinois at Urbana-Champaign, USA.

* Corresponding author.

E-mail addresses: qntran2@illinois.edu (Q. N. V. Tran), jroesler@illinois.edu (J. R. Roesler).

## ABTRACT

Joint raveled or spalled damage (henceforth called joint damage) can affect the safety and long-term performance of concrete pavements. It is important to assess and quantify the joint damage over time to assist in building action plans for maintenance, predicting maintenance costs, and maximize the concrete pavement service life. A framework for the accurate, autonomous, and rapid quantification of joint damage with a low-cost camera is proposed using a computer vision technique with a deep learning (DL) algorithm. The DL model is employed to train 263 images of sawcuts with joint damage. The trained DL model is used for pixel-wise color-masking joint damage in a series of query 2D images, which are used to reconstruct a 3D image using open-source structure from motion algorithm. Another damage quantification algorithm using a color threshold is applied to detect and compute the surface area of the damage in the 3D reconstructed image. The effectiveness of the framework was validated through inspecting joint damage at four transverse contraction joints in Illinois, USA, including three acceptable joints and one unacceptable joint by visual inspection. The results show the framework achieves 76% recall and 10% error.

## 1. Introduction

Sawcutting concrete pavements is an important step in controlling the location of early-age shrinkage cracks [1,2]. Likewise, minimizing the raveling or spalling (henceforth, called joint damage) during the sawcutting operation requires the proper timing and equipment. The most commonly used method for initiating sawing is to scratch the slab surface with a penknife or nail, or to stand on the slab to observe footprint impression [2]. The extent of raveling during the initial sawcutting provides the operator with visual information on whether it is the proper time to continue or wait for further concrete maturity (hydration). This feedback loop employed by sawcutting personnel is experiential and subjective [3], which leads to occasional projects with excessive joint raveling or delayed sawcuts and premature cracking.

At later ages of concrete maturity, joint spalling damage can be caused by the contribution of traffic loads, weather conditions, incompressible, and concrete quality [4,5]. Furthermore, other durability distress mechanisms such as calcium oxychloride formation can produce joint spalling and damage [6,7]. Whether the joint damage is caused by raveling at early ages or later age spalling, it negatively affects vehicle safety, ride quality, and overall performance life of the concrete pavement [8-10]. If joint damage is not repaired properly and promptly, repair cost increase significantly with further delays [9]. Particularly for airfield pavements, joint damage can lead to Foreign Object Debris (FOD) [11], which costs the aerospace industry $4 billion annually [8]. As such, pavement engineers must frequently assess and measure the joint quality of the concrete pavement for the safe operation of aircraft. However, the existing inspection methodology for joint damage is visually conducted by an experienced engineer in order to calculate a gross Pavement Condition Index (PCI) [12], which is described in detail in the Test Method for Airport Pavement Condition Index Survey [13]. The joint damage for this method is estimated based on the severity and extent and is combined with other distress data (e.g., cracking, to define the current PCI. Unfortunately, this method is slow, done on annual or bi-annual basis, and does not specifically quantify the joint damage over time but the over performance of the pavement over time.

To improve estimating joint raveling level and joint damage development over time, an autonomous and reliable measurement method is needed for quantifying joint damage by field construction personnel and pavement engineers. Additionally, joint damage measurement over time will assist in developing better maintenance and rehabilitation plans.

## 1.1. Current methods to assess concrete joint conditions

There are multiple methods to quantify joint damage with almost all of them falling into qualitative judgments of extent and severity for a particular survey location. The Federal Highway Administration (FHWA) manual introduces a method to classify spalling in three distinct severity levels (low, moderate, and high) by measuring the width of the spall [14]. The PCI discussed above is a numeric score from 0 (failed) to 100 (good) determined from a visual rating of the distress condition of roads and airfields that includes joint damage as one of its many factors [12]. The joint damage index (JRI) was proposed by Krstulovich et al. [1] to visually rate contraction joint raveling especially with the application of early entry saws. JRI is a numerical rating of observed joint raveling damage using a relative comparison, where 0 and 5 represent the least and most extensive damage, respectively. In recent years, pavement surface images and videos have been collected by high-speed digital inspection vehicles and later reviewed by engineers to manually rate and assess observed defects including joint damage [11]. Although digital inspections have improved the speed the pavement evaluation data is collected as well as the safety of the raters, it is still a time-consuming and costly task to manually rate videos/images and the final results of joint damage are influenced by the subjectivity and experience of the evaluators [15].

In recent years, automated image analyses have been introduced to quantify certain distress collected from high-speed pavement evaluation data. Arena et al. [16] proposed a crack quantification method based on 2D image analysis with application to microcrack description in rocks. Liu et al. [17] developed a crack image analysis system using 2D images and applied the system for soil crack patterns and rock features. In comparison, the adaptation of machine learning and deep learning-based approaches for automated pavement distress detection has shown higher accuracy and shorter computational time demand compared to 2D image detection techniques [18-20]. Dorafshan et al. [21] demonstrated that the supervised deep convolution neural network (CNN) in transfer learning mode was able to detect 86% of cracked images with crack widths larger than 0.04 mm while the edge detection methods were able to detect 53-79% of cracked pixels with cracks wider than 0.1 mm.

Researchers have begun using three-dimensional (3D) images from laser scanning [22,23] or 3D reconstructed images using structure from motion [24,25]. Zhang [23] proposed CrackNet for pixel-level crack detection using a dataset of 2,000 3D images of asphalt surfaces. This method is expensive and large 1-mm 3D image dataset for training is neither available nor easy to be collected. Torok et al. [25] developed a crack detection algorithm (CDA) to assess and analyze the surface damage of post-disaster buildings based on a 3D point cloud image reconstructed using structure from motion [24]. However, manipulating the orientation of 3D reconstructed image and manual selecting angle threshold are time-consuming and reduces automaticity of the method. In addition, surface tining, macrofibers, texture, and sawcut grooves cause false detection and reduces the accuracy of the damage quantification algorithm.

## 1.2. Research objective

This chapter proposes a 3D damage quantification algorithm (DQA) with a low-cost portable camera (e.g., a phone camera) for automatic quantification of joint damage by combining 3D reconstruction image with a deep learning (DL) model. The DL model is used for detecting the damage regions from the image background features (e.g., tining, texture, fibers, and sawcut groove). The 2D images output from the DL model are used to construct a 3D image. The DQA algorithm analyzes the 3D image to localize the damage using red color thresholding and calculates a joint damage index (JDI) that quantitatively defines the joint damage extent and severity. The proposed DQA and JDI provides the saw operators, pavement evaluators, and forensic investigators an effective quantitative measurement tool to support: (1) sawcut initiation decisions; (2) verify acceptable joint raveling level after sawcutting; (3) quantification of joint spalling; and (4) a time history of joint damage for same facility to build more effective maintenance and rehabilitation plans.

## 2. Methodology

Automatic quantification of the extent of joint damage is important for concrete pavements. Feeding unnecessary information from an image such as the background and other surface features (e.g., tining, sawcut groove, fibers, and texturing) will prolong the processing time by increasing 3D image reconstruction time as well as reduce its accuracy. To improve automation and accuracy of damage detection and quantification, a DL model [26] is embedded in DQA as a pre-processing step for autonomously detecting contraction joints with varying levels of damage. A proposed hierarchical approach is shown in Fig. 1. The approach has three main stages including:

- Stage 1: Train DL model to detect the joint damage in 2D images which contain unnecessary background information;
- Stage 2: Implement the post-trained DL model on query or new 2D images to output 2D images containing detected joint damage areas;
- Stage 3: Reconstruct a 3D image using the 2D images in stage 2. Finally, a quantification algorithm localizes the 3D joint damage and computes the damage surface area and JDI.

The following sections describe each step in details and contributions to this new approach.

## 2.1. Stage 1: Training deep learning model

### 2.1.1. Mask R-CNN architecture

The first step in the overall framework is to train a model for pixel-level detection of joint damage. A DL model, Mask R-CNN using the backbone of ResNet-101 [27] and feature pyramid network (FPN) [28], was chosen to train a dataset of concrete joint images because it is state-of-the-art in pixel-wise object detection [26]. Fig. 2 shows the schematic of Mask R-CNN. It is proven to be a simple and fast system and it outperformed previous single model entries on every task in the 2016 COCO challenge [29]. As shown in Fig. 2, a concrete input image is fed into a convolution neural network (CNN) layer with ResNet-101, which is really a DL network with 101 layers and FPN backbone for feature extraction. The region proposal network takes feature map output from the last shared CNN layer as input, slides a small network over the feature map, and outputs the Region of Interest (RoI). RoI-Align properly aligns the extracted features with the input image pixel-to-pixel and feeds into the new convolution layer for instance segmentation, outputting a binary mask layer indicating joint damage locations (see Fig. 5).

### 2.1.2. Data collection and augmentation

A dataset containing 263 images of concrete pavement contraction joints has been collected in a diverse set of lighting and weather conditions in Champaign-Urbana, IL (USA) using a low-cost camera as the RGB data sensor. Each image has a maximum size of 1024×1920 pixels. Image annotation is completed by VGG VIA [30]. For the 263 image dataset, 90% (235 images) of the images are used for training and 10% (28 images) for validation.

A common strategy to train a DL model is collecting a sufficiently large dataset. However, when a dataset for training is small, an alternative strategy is to enlarge the size of the dataset through artificial image augmentation. By applying the image augmentation, the DL model is trained on how to recover from a poorer positioning or orientation of the joint in the image, providing good performance, improving generalization, and reducing overfitting [31-34]. In this study, multiple augmentation techniques were applied randomly on the training images to expand the dataset. Fig. 3 demonstrates an example training image augmented by the different augmentation techniques, such as Gaussian blurring, brightening or darkening images, vertical or horizontal flipping. These image augmentations significantly increased the size of the training set from 263 to 1578 images.

### 2.1.3.   Training configuration

Transfer learning is applied during training as it improves learning in a new task through the transfer of knowledge from a related task that has already been learned [35]. There are two common types of transfer learning processes:  pre-training and fine-tuning. Due to our small dataset and the similarity of sawcut images, pre-training is implemented initially using the COCO weights obtained from pre-training on the Microsoft COCO dataset [29]. The pre-training consists of three stages (see Fig. 4). In the first stage, only the prediction heads or fully connected (FC) layers are trained with a learning rate of 0.001. The validation loss initially reduces significantly and levels off after 182 epochs. After the first training, validation loss improves by training the ResNet-101 network stage 4 and up for another 200 epochs with the same learning rate. Finally, the learning rate is reduced by a factor of 10 for the entire model training (fine-tuning) for an additional 100 epochs as seen in Fig. 4.4. The training optimization is performed using a stochastic gradient descent (SGD) with weight decay set to 0.0001.

The DL model is trained using a workstation with an Intel Core i7-7700 @ 3.6GHz CPU, 32GB Ram, and NVIDIA 1080 Ti GPU. It took approximately 12 hours for training the model until the validation loss was convergent (e.g., 1.29). The trained weights were saved for pixel-wise damage detection of query or new joint images.

## 2.2.   Stage 2: Color masking of joint damage in 2D images

From a recorded video of a 500 mm portion of a sawcut joint, more than 50 high-resolution images (1080 × 1920 pixels) were extracted in order to have sufficient overlap between adjacent images for feature matching during 3D image reconstruction. The images were input into the post-trained DL model to output a binary mask image indicating pixels with joint damage as shown in Fig. 5. The identifying pixels encompassing the damage area were set to an arbitrary RGB value (e.g., RGB = 255, 0, 0).

## 2.3.   Stage 3: Quantification of joint damage using 3D reconstructed image

With the input of the color-masked 2D images, a 3D image was reconstructed from the sequence of the images using an open-source image-based 3D reconstruction solution, AliceVision [36]. The 3D reconstruction solution used scale-invariant feature transform (SIFT) [37] for key feature matching and structure from motion steps. The output 3D reconstructed image contains the damages color-masked in 2D images (see second image from the left in Fig. **10** and Fig. 11). However, the red color intensity of 3D damage areas does not remain at 255 as in the input 2D color-masked images due to the process of averaging pixel colors [38-40].

In order to localize the damaged region in the 3D image, the DQA applies red color thresholding to the 3D image using. The algorithm counts triangular faces in the 3D image as joint damage when the red color intensity of three vertices of a face is higher than a given red color threshold and sets the face to a new RGB = (0, 255, 0) as indicated under green color. After the joint damage domain (D) are detected, the surface area ($S_D^i$) of each damage element or face $i^{th}$ is computed. Finally, to allow for a comparison of joint damage of different sawcut lengths, the joint damage surface area is normalized to a fixed projected surface area to define the joint damage index (JDI) (see Eq. 4.1). The fixed surface area for the JDI denominator is defined as the product of the observed sawcut length (500 mm) and 3 times the maximum aggregate size ($D_{max}$=25 mm) [41].

$$JDI(\%) = \frac{\sum_D S_D^i}{3 \times 500 \times D_{max}}$$ (Eq. 1)

## 3. Experimental results and discussion

The proposed framework was field implemented on four completely new 500-mm long transverse contraction joints belonging to separate construction sites. The joints are video captured under various weather conditions: a sidewalk (SW) on Wright Street in Champaign, IL, two parking lots (Parking lot #1 and #2) in Urbana, IL, and a tollway jointed plain concrete pavement (JPCP) near Itasca, IL (USA). The concrete sections were selected because they had various surface textures and appearances. Specifically, the sidewalk has a smooth surface, the parking lots showed a higher textured surface with macrofibers and slight scaling, and the JPCP section had longitudinal tining and texture. All of these various surface features act as noise and increase the difficulty in detecting the joint damage accurately. The sawn joints introduced in Fig. 8.7 consist of a variety of joint damage sizes with the smallest length approximately 13 mm. Of the four contraction joints, only one is rated visually as unacceptable while the other three are acceptable. A 12MP smartphone camera (e.g., iPhone 7) was used to video record the sawcuts to increase the number of images at different viewpoints. The next sub-sections explain the 2D and 3D damage analysis and the effect of color threshold on joint damage quantification.

### 3.1. Performance of joint damage detection in 2D images

The performance of 2D damage detection is examined based on (i) recall and (ii) error metrics from common semantic segmentation and scene parsing evaluation [42,43]. Recall or accuracy measures the proportion of real positives ($GT_d$) that are the correctly predicted positive (TP) (see Fig. 6). The error measures the false predicted positive cases (FP) over real positives

GT$_d$ and it is essential to avoid overestimating joint damage. The recall and error formulas are shown below.

$$recall = \frac{TP}{TP + FN} = \frac{TP}{GT_d} \qquad \text{(Eq. 1)}$$

$$error = \frac{FP}{TP + FN} = \frac{FP}{GT_d} \qquad \text{(Eq. 2)}$$

where TP and FP (see Fig. 6) are the area of the correctly and false predicted regions, respectively.

The post-trained DL model detected joint damages in the 2D images of the sawcuts shown in Fig. 8. To examine the performance of damage detection in 2D images, 10% of 2D images from the field sections were selected randomly and manually annotated using VIA VGG annotation software [30]. The mean recall and error for 2D damage segmentation results are shown in Table *1*. The results show that the recall of damage detection of the post-trained model is high but so is the error. This indicates the model was able to detect the true joint damages but also it overestimated its extent. In addition, Fig. 9 shows examples of high false detections in 2D images by the post-trained DL model.

3.2. Performance of 3D images and effect of color threshold on damage quantification

Fig. 10 and Fig. 11 introduces the 3D reconstructed images of the four sawcut joints with the red color indicating the spalling damage (first image on the left), the image containing the ground truth damage (yellow), and next four images showing damage detected by color thresholds (green) of 190, 210, 230, and 250, respectively. Similar to evaluating the performance of damage detection in 2D images, the performance of damage quantification in 3D images is examined by recall and error in equations 4-2 and 4-3. The recall, error, and JDI for the four sawn joints are summarized in Table 2. To calculate the ground truth, the damage (indicated as green yellow) is segmented out manually from the 3D images and compute JDI by summing all the surface area within the segmented damage.

Fig. 12 shows the relationship between the recall and error of the damage detections for a range of color thresholds. The highest and lowest recall of 93% and 29% correspond to the color threshold of 190 and 250, respectively. Color thresholds equal to or less than 230 provide comparable recall for the four sites. The highest and lowest error values of 135% and almost 0% correspond to the color thresholds of 190 and 250, respectively. For all field sites, low error is for thresholds of 230 and 250. Based on the fact that the JDI of sawn joints accepted by pavement engineers is less than 3% [41], the JDI results at the thresholds 190 and 210 especially for PL1 and JPCP are not in good agreement with the visual inspection. The color threshold of 230 was chosen as a balance between recall and error and provides the JDI in good agreement with the observed joint damage of all sawn joints on each of the 4 project sites.

Comparing the performance of joint damage detection with 2D image, using the red color threshold of 230 with 3D reconstructed images reduced the recall slightly from 82% to 76% but improved the error significantly from 120% to 10%.

### 3.3. Computation time

The amount of processing time for the entire 3D damage quantification process including each step is summarized in Table 3. The initial 12 hours for training the DL model was excluded. The total processing time varied from 33 to 64 minutes mainly from the 3D image reconstruction step. Meshing from a dense point cloud is a time-consuming step in the 3D reconstruction and it is affected by many parameters [44]. Pixel-wise detecting and color-masking joint damage in 2D images take 3.3 seconds per image on average (see Fig. 13) and 3D damage recognition and quantification takes less than a minute using Intel Core i7-7700 @ 3.6GHz CPU with 32GB Ram and NVIDIA 1080 Ti GPU.

## 4. Conclusions

Knowledge of the joint damage occurring on concrete pavement contraction joints can improve construction quality and monitor joint performance over time more accurately for planning maintenance and rehabilitation schedules. This chapter proposed a 3D damage quantification algorithm (DQA) using a CV technique with DL model for accurate and automatic quantification of joint damage. The DL model was trained on 263 images for 12 hours and used for pixel-wise damage detection in 2D images, which were subsequently used to reconstruct a 3D image. The proposed DQA applied a red color threshold to localized and quantify the damage in the 3D reconstructed image. To validate the technique, field tests were conducted on four transverse contraction joints of different types of pavements, such as a sidewalk, parking lots, and JPCP. The joint damages along those contraction joints have various sizes and lengths, the minimum length is approximately 13 mm. The red color threshold of 230 was found to be optimal for joint damage quantification. With a total of 263 images used for training the DL model, the results show that the technique quantified the joint damage with 76% recall and 10% error. The total computation time varied from 13 and 64 minutes, mainly because of reconstructing 3D images. The pixel-wise damage detection takes 3 seconds per image and DQA takes 6 seconds on average. The recall of the framework can still be improved when more images of joint damage continue to be added for retraining the DL model. Additionally, this technique can quantify 3D joint damage without manual coordinate manipulation or gravity alignment, eliminating a time-consuming step in 3D image processing. The automatic framework can provide a standardized procedure for QC/QA for personnel with a variety of experience to assess and monitor the quality of a sawcut for in situ concrete pavements and slabs for sawcut commencement timing or maintenance monitoring. Furthermore, the framework can be implemented in mobile platforms to quantify and analyze pavement distress.

## 5. Acknowledgement

This research did not receive any specific grant from funding agencies in the public, commercial, or not-for-profit sectors.

## 6. References

[1]     J. Krstulovich Jr, T. Van Dam, K. Smith, M. Gawedzinski, Evaluation of potential long-term durability of joints cut with early-entry saws on rigid pavements, Transportation Research Record: Journal of the Transportation Research Board (2235) (2011) 103-112.

[2]     P. Okamoto, P. Nussbaum, K. Smith, M. Darter, T. Wilson, C. Wu, S. Tayabji, Guidelines for timing contraction joint sawing and earliest loading for concrete pavements. Volume II, Report No. FHWA-RD-91-080, Federal Highway Administration, Mclean, VA, 1994.

[3]     P. Taylor, X. Wang, Concrete Pavement Mixture Design and Analysis (MDA): Comparison of Setting Time Measured Using Ultrasonic Wave Propagation with Saw-Cutting Times on Pavements in Iowa, Report No. TPF-5 (205), National Concrete Pavement Technology Center, Ames, IA, 2014.

[4]     L. Wang, D.G. Zollinger, Mechanistic design framework for spalling distress, Transportation Research Record: Journal of the Transportation Research Board 1730 (1) (2000) 18-24.

[5]     D.G. Zollinger, S.P. Senadheera, T. Tang, Spalling of continuously reinforced concrete pavements, Journal of Transportation Engineering 120 (3) (1994) 394-411.

[6]     J. Monical, E. Unal, T. Barrett, Y. Farnam, W.J. Weiss, Reducing joint damage in concrete pavements: Quantifying calcium oxychloride formation, Transportation Research Record: Journal of the Transportation Research Board 2577 (1) (2016) 17-24.

[7]     J. Weiss, M.T. Ley, L. Sutter, D. Harrington, J. Gross, S.L. Tritsch, Guide to the Prevention and Restoration of Early Joint Deterioration in Concrete Pavements, 2016.

[8]     J. Patterson Jr, Foreign object debris (FOD) detection research, International Airport Review 11 (2) (2008) 22-27.

[9]     T.J. Freeman, J.D. Borowiec, B. Wilson, P. Arabali, M. Sakhaeifar, Pavement maintenance guidelines for general aviation airport management, 2016.

[10]    J. Greene, M. Shahin, D. Alexander, Airfield pavement condition assessment, Transportation Research Record: Journal of the Transportation Research Board (1889) (2004) 63-70.

[11]    C. Koch, I. Brilakis, Pothole detection in asphalt pavement images, Advanced Engineering Informatics 25 (3) (2011) 507-515.

[12]    M.Y. Shahin, Pavement management for airports, roads, and parking lots, Springer New York, 2005.

[13]    ASTM D5340-12, Standard Test Method for Airport Pavement Condition Index Surveys, 2010.

[14]    J.S. Miller, W.Y. Bellinger, Distress identification manual for the long-term pavement performance program, United States. Federal Highway Administration, 2014.

[15]    A. Bianchini, P. Bandini, D.W. Smith, Interrater reliability of manual pavement distress evaluations, Journal of Transportation Engineering 136 (2) (2010) 165-172.

[16]    A. Arena, C. Delle Piane, J. Sarout, A new computational approach to cracks quantification from 2D image analysis: Application to micro-cracks description in rocks, Computers & Geosciences 66 (2014) 106-120.

[17]    C. Liu, C.-S. Tang, B. Shi, W.-B. Suo, Automatic quantification of crack patterns by image processing, Computers & Geosciences 57 (2013) 77-80.

[18]    Y.J. Cha, W. Choi, O. Büyüköztürk, Deep learning-based crack damage detection using convolutional neural networks, Computer-Aided Civil and Infrastructure Engineering 32 (5) (2017) 361-378.

[19]    L. Some, Automatic image-based road crack detection methods, 2016.

[20]    D. Xie, L. Zhang, L. Bai, Deep Learning in Visual Computing and Signal Processing, Applied Computational Intelligence and Soft Computing 2017 (2017) 13.

[21]    S. Dorafshan, R.J. Thomas, M. Maguire, Comparison of deep convolutional neural networks and edge detectors for image-based crack detection in concrete, Construction and Building Materials 186 (2018) 1031-1045.

[22]    Q. Li, M. Yao, X. Yao, B. Xu, A real-time 3D scanning system for pavement distortion inspection, Measurement Science and Technology 21 (1) (2009) 015702.

[23]    A. Zhang, K.C. Wang, B. Li, E. Yang, X. Dai, Y. Peng, Y. Fei, Y. Liu, J.Q. Li, C. Chen, Automated pixel-level pavement crack detection on 3D asphalt surfaces using a deep-learning network, Computer-Aided Civil and Infrastructure Engineering 32 (10) (2017) 805-819.

[24]    C. Wu, VisualSFM: A visual structure from motion system, (2011).

[25]    M.M. Torok, M. Golparvar-Fard, K.B. Kochersberger, Image-based automated 3D crack detection for post-disaster building assessment, Journal of Computing in Civil Engineering 28 (5) (2013) A4014004.

[26]    K. He, G. Gkioxari, P. Dollár, R. Girshick, Mask r-cnn, Computer Vision (ICCV), 2017 IEEE International Conference on, IEEE, 2017, pp. 2980-2988.

[27]    K. He, X. Zhang, S. Ren, J. Sun, Deep residual learning for image recognition, Proceedings of the IEEE conference on computer vision and pattern recognition, 2016, pp. 770-778.

[28]    T.-Y. Lin, P. Dollár, R.B. Girshick, K. He, B. Hariharan, S.J. Belongie, Feature Pyramid Networks for Object Detection, CVPR, Vol. 1, 2017, p. 4.

[29]    T.-Y. Lin, M. Maire, S. Belongie, J. Hays, P. Perona, D. Ramanan, P. Dollár, C.L. Zitnick, Microsoft coco: Common objects in context, European conference on computer vision, Springer, 2014, pp. 740-755.

[30]    A. Dutta, A. Gupta, A. Zissermann, VGG Image Annotator (VIA), URL: http://www. robots. ox. ac. uk/~ vgg/software/via (2016).

[31]    J.W. Johnson, Adapting Mask-RCNN for Automatic Nucleus Segmentation, arXiv preprint arXiv:1805.00500 (2018).

[32]    A. Krizhevsky, I. Sutskever, G.E. Hinton, Imagenet classification with deep convolutional neural networks, Advances in neural information processing systems, 2012, pp. 1097-1105.

[33]    X. Cui, V. Goel, B. Kingsbury, Data augmentation for deep neural network acoustic modeling, IEEE/ACM Transactions on Audio, Speech and Language Processing (TASLP) 23 (9) (2015) 1469-1477.

[34]    L. Perez, J. Wang, The effectiveness of data augmentation in image classification using deep learning, arXiv preprint arXiv:1712.04621 (2017).

[35] L. Torrey, J. Shavlik, Transfer learning, Handbook of research on machine learning applications and trends: algorithms, methods, and techniques, IGI Global, 2010, pp. 242-264.

[36] P. Moulon, P. Monasse, R. Marlet, Adaptive structure from motion with a contrario model estimation, Asian Conference on Computer Vision, Springer, 2012, pp. 257-270.

[37] D.G. Lowe, Distinctive image features from scale-invariant keypoints, International journal of computer vision 60 (2) (2004) 91-110.

[38] B. Lévy, S. Petitjean, N. Ray, J. Maillot, Least squares conformal maps for automatic texture atlas generation, ACM Transactions on Graphics (TOG) 21 (3) (2002) 362-371.

[39] H.-Y. Shum, R. Szeliski, Panoramic image mosaics, Citeseer, 1997.

[40] Y. Sato, M.D. Wheeler, K. Ikeuchi, Object shape and reflectance modeling from observation, (1997).

[41] Q. Tran, R. Jeffery, Non-Contact Ultrasonic and Computer Vision Assessment for Sawcut Initiation, Journal of Transportation Engineering, Part B: Pavements (submitted 2019).

[42] J. Long, E. Shelhamer, T. Darrell, Fully convolutional networks for semantic segmentation, Proceedings of the IEEE conference on computer vision and pattern recognition, 2015, pp. 3431-3440.

[43] D.M. Powers, Evaluation: from precision, recall and F-measure to ROC, informedness, markedness and correlation, (2011).

[44] A. Maiti, D. Chakravarty, Performance analysis of different surface reconstruction algorithms for 3D reconstruction of outdoor objects from their digital images, SpringerPlus 5 (1) (2016) 932.

## 7. Tables

**Table 1** – Average recall and error segmentation of joint damage in 2D images.

| Metrics | Sidewalk | Parking Lot#1 | Parking Lot#12 | JPCP | Overall average |
|---|---|---|---|---|---|
| Mean recall | 90% | 80% | 83% | 76% | 82% |
| Mean error | 235% | 53% | 72% | 72% | 120% |

**Table 2**– Recall, error, and JDI by the 3D damage quantification for several red color thresholds.

| | Sidewalk (SW) | Parking lot #1 (PL1) | Parking lot #2 (PL2) | Concrete Pavement (JPCP) | Average |
|---|---|---|---|---|---|
| **Visual inspection** | Acceptable | Acceptable | Un- acceptable | Acceptable | |
| Color threshold 190 | | | | | |
| Recall | 93% | 87% | 87% | 91% | 90% |
| Error | 135% | 33% | 347% | 119% | 158% |
| JDI | 3% | 4% | 95% | 7% | |
| Color threshold 210 | | | | | |
| Recall | 85% | 84% | 82% | 85% | 84% |
| Error | 38% | 12% | 97% | 33% | 45% |
| JDI | 2% | 4% | 39% | 4% | |
| Color threshold 230 | | | | | |
| Recall | 75% | 78% | 72% | 80% | 76% |
| Error | 7% | 6% | 13% | 14% | 10% |
| JDI | 1.5% | 3% | 19% | 3% | |
| Color threshold 250 | | | | | |
| Recall | 54% | 60% | 29% | 64% | 51% |
| Error | 1% | 2% | 0% | 6% | 2% |
| JDI | 1% | 2% | 6% | 2% | |

**Table 3 –** Processing time for 3D joint damage quantification for joints on four construction sites.

| Steps | Total computation time (minutes) | | | |
|---|---|---|---|---|
| | **Sidewalk (SW)** | **Parking lot #1 (PL1)** | **Parking lot #2 (PL2)** | **Concrete Pavement (JPCP)** |
| Images | 47 | 46 | 50 | 76 |
| Detecting and masking damage | 3 | 3 | 2 | 4 |
| 3D image reconstruction | 30 | 38 | 11 | 60 |
| 3D damage quantification | <1 | <1 | <1 | <1 |
| Total | 34 | 42 | 13 | 64 |

## 8. Figures

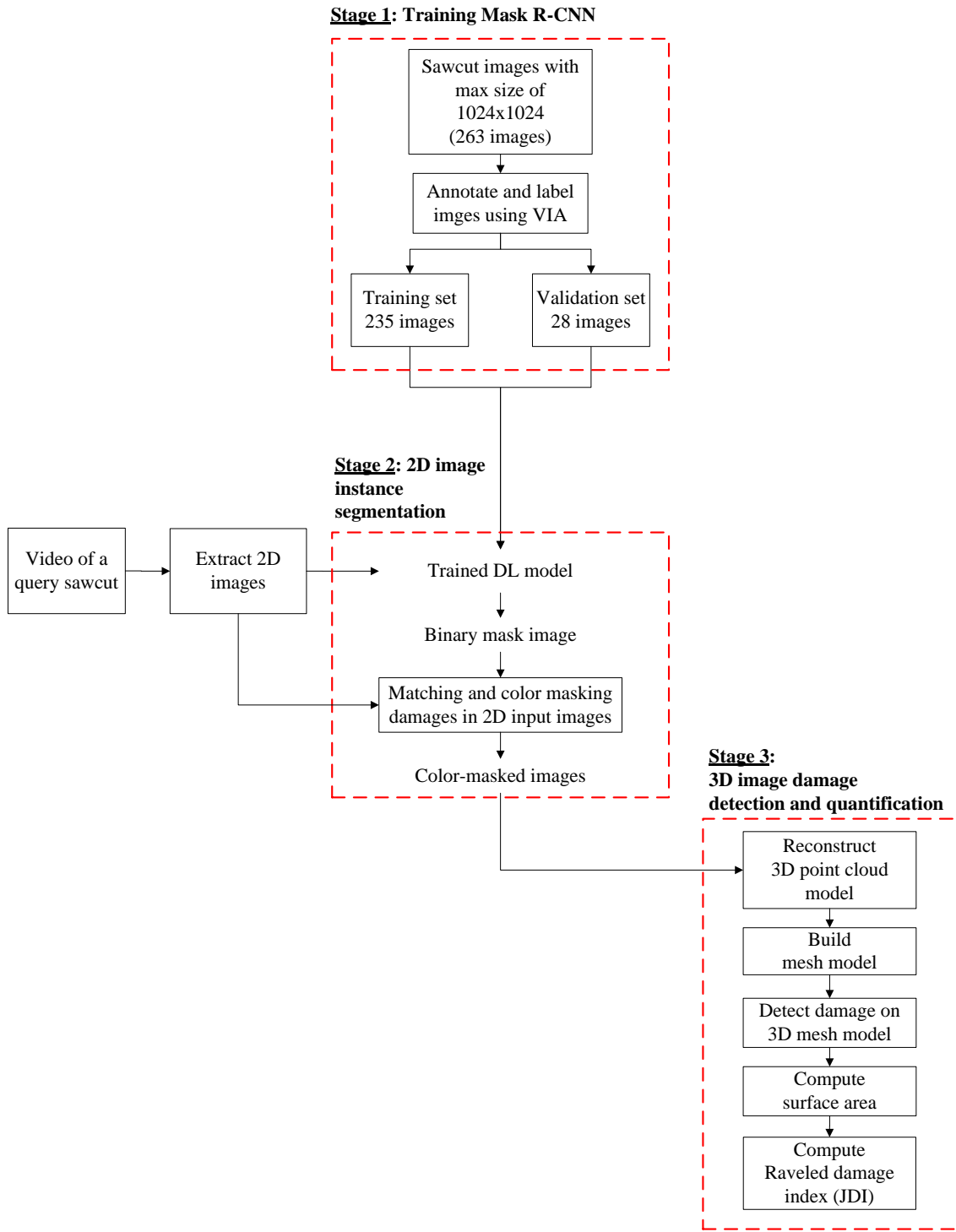

**Fig. 1** – Overall framework of the DL model for joint raveling and spalling damage detection and quantification.

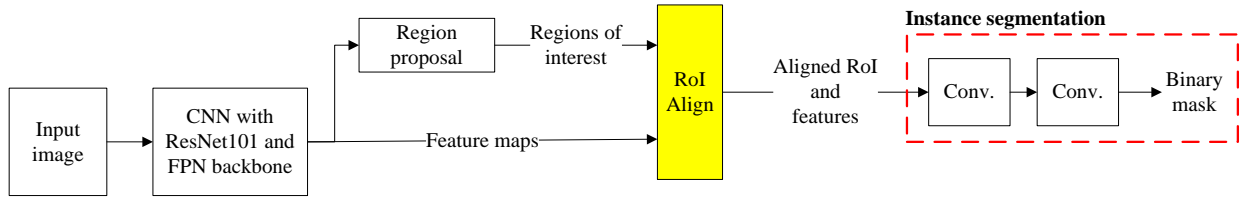

**Fig. 2** – Schematic of Mask R-CNN using FPN-ResNet101 backbone for 2D damage segmentation [26].

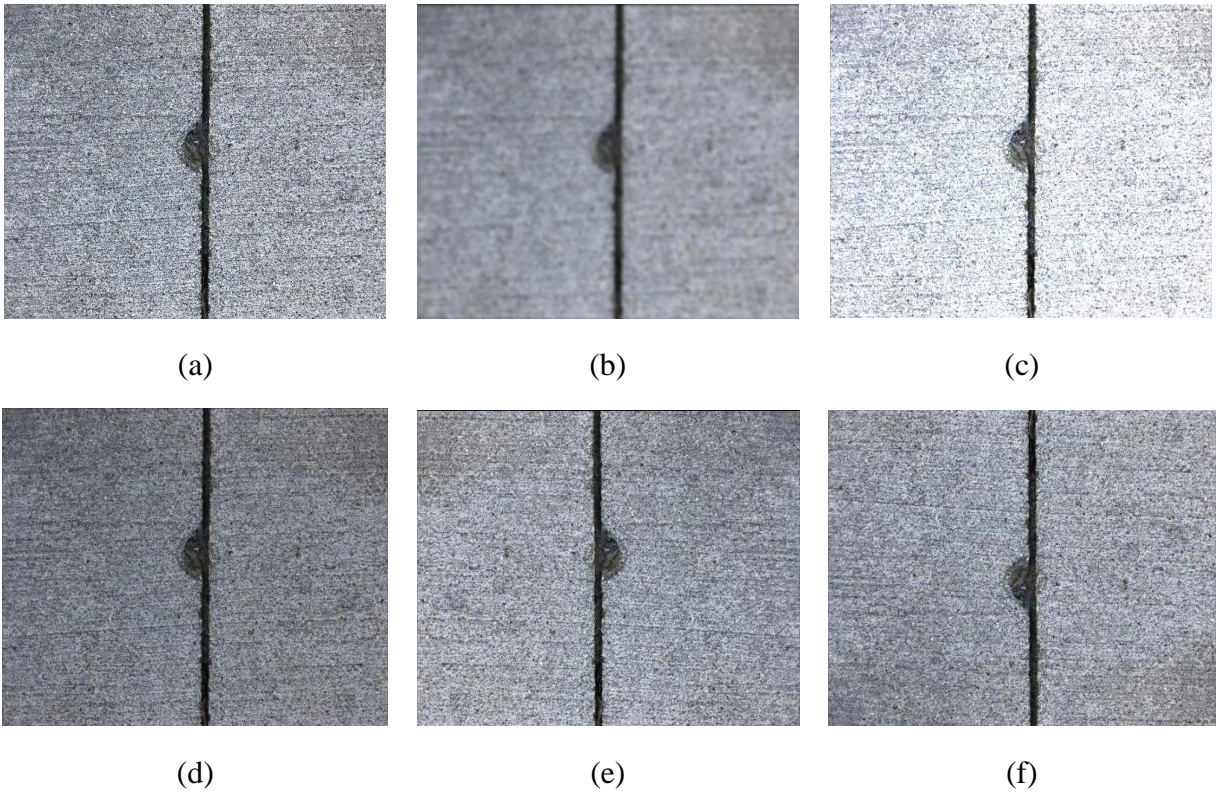

**Fig. 3** – Augmentation methods: (a) original image (none), (b) Gaussian blurring with a sigma of 0.25, (c) brightening, (d) darkening, (e) vertical flip, and (f) horizontal flip.

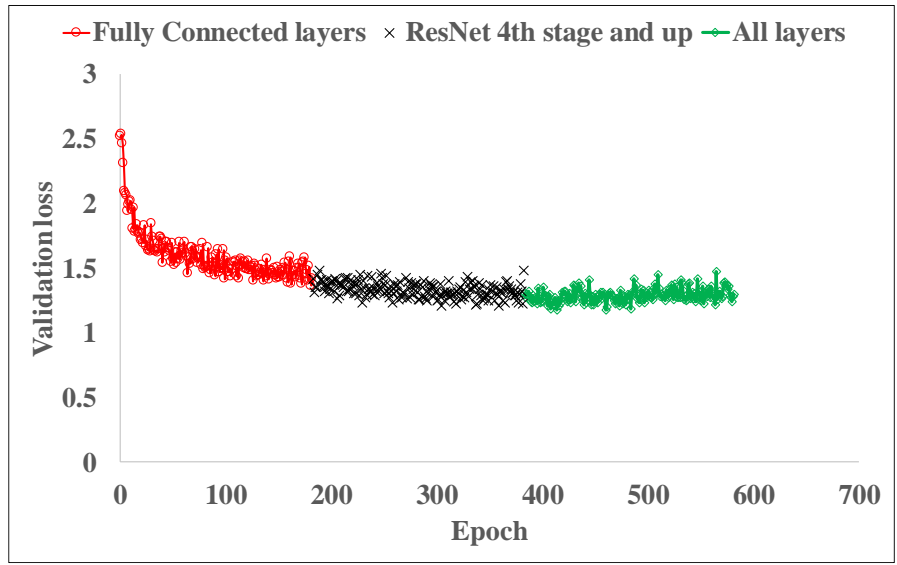

**Fig. 4** – Traces of validation total loss during training the DL model with initial COCO pre-trained weights *[29]* for joint damage instance segmentation.

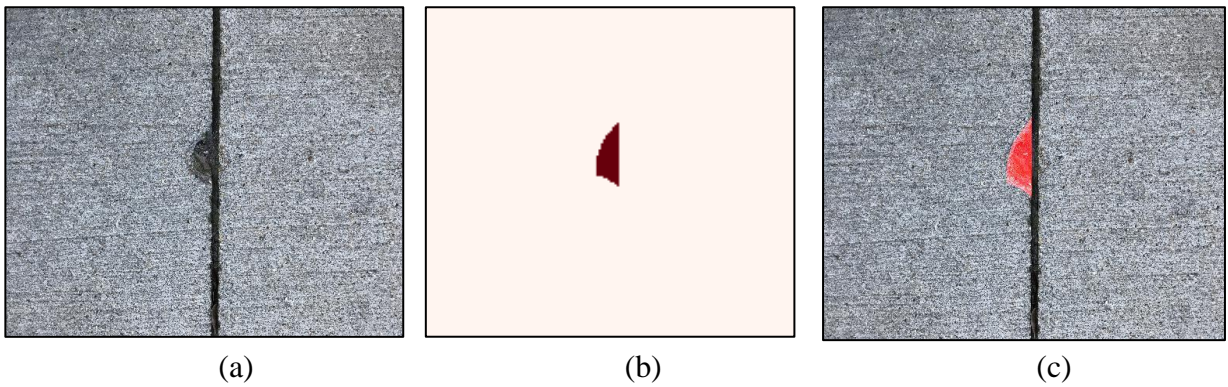

<div align="center">(a)        (b)        (c)</div>

**Fig. 5** – (a) Input image, (b) binary mask image, and (c) color-masked image of joint damage.

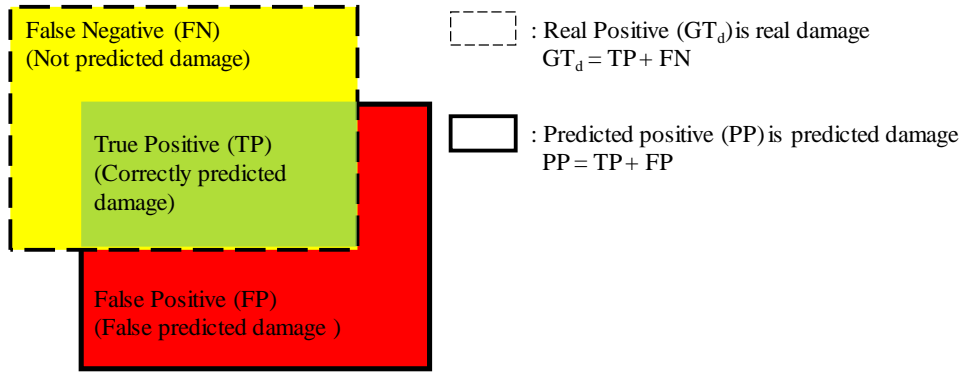

**Fig. 6** – Illustration showing ground truth GT$_d$ (real damage or positive), false positive FP (incorrectly predicted damage) in red, false negative FN (not predicted damage) in yellow, and true positve TP (correct predicted damage) in green, which is part of the real damage.

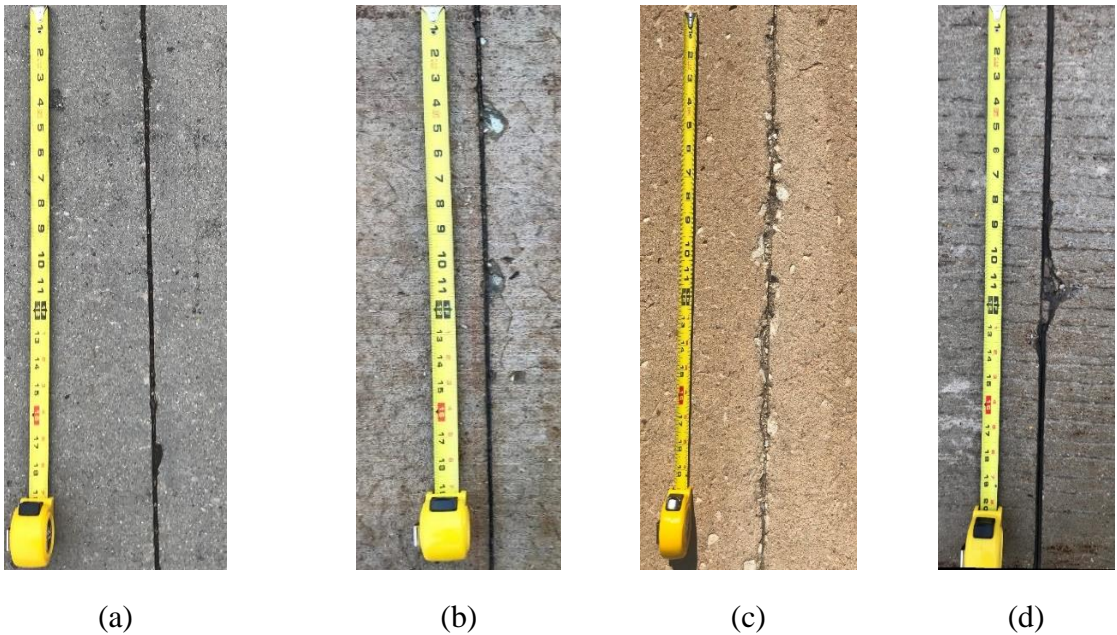

|  (a) | (b) | (c) | (d) |

**Fig. 8.7:** Overview images of four sawcut contraction joints (a) sidewalk SW, (b) parking lot#1, (c) parking lot#2, and (d) JPCP captured under different weather conditions. The pavement surfaces have various surface textures and appearances.

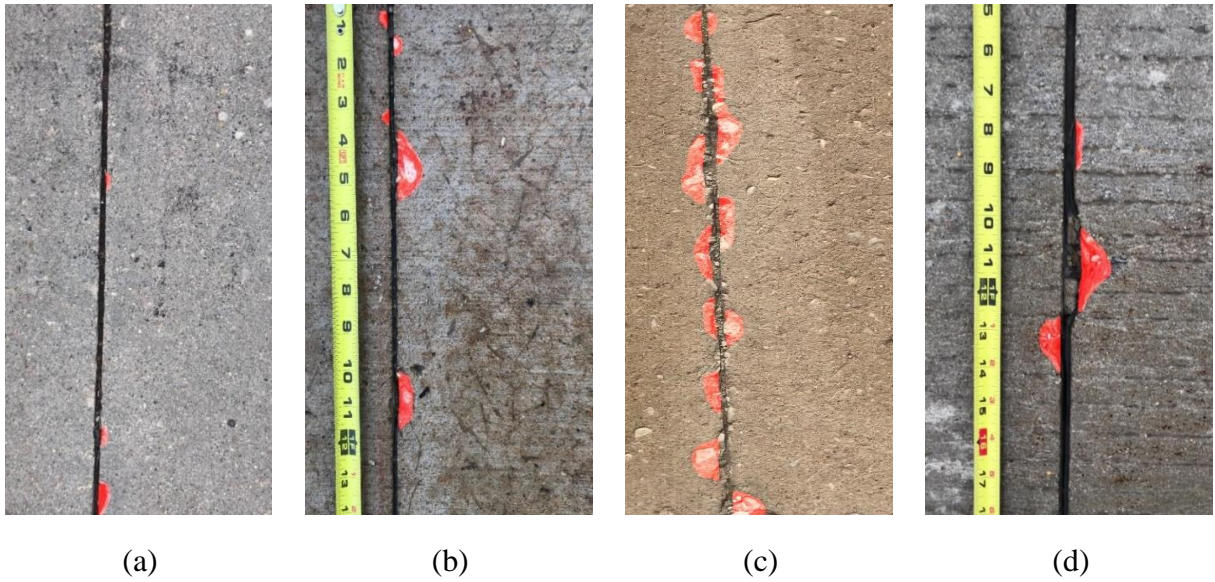

(a)  (b)  (c)  (d)

**Fig. 8** – Examples of color-masked 2D images at contraction joints at four project locations: (a) sidewalk, (b) parking lot#1, (c) parking lot#2, and (d) JPCP.

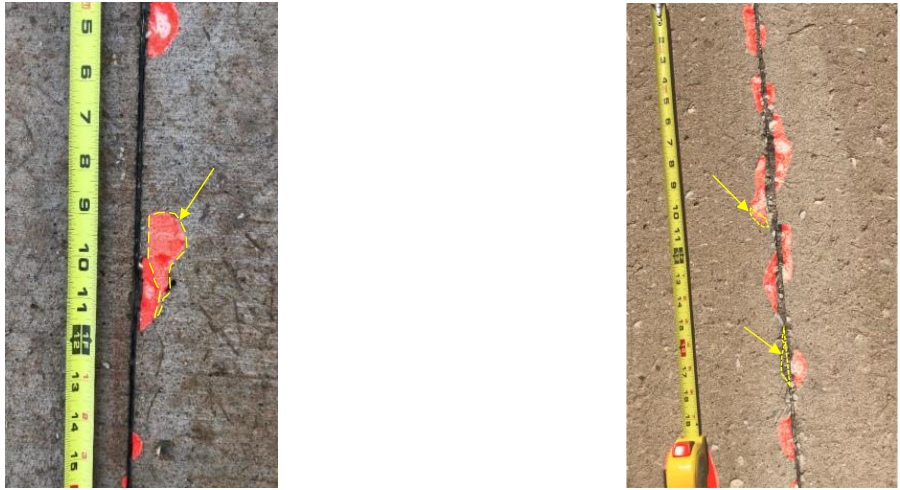

**Fig. 9** – Typical false positives (area bounded by the dashed yellow line) by trained DL at joints on (a) parking lot#1 and (b) parking lot#2.

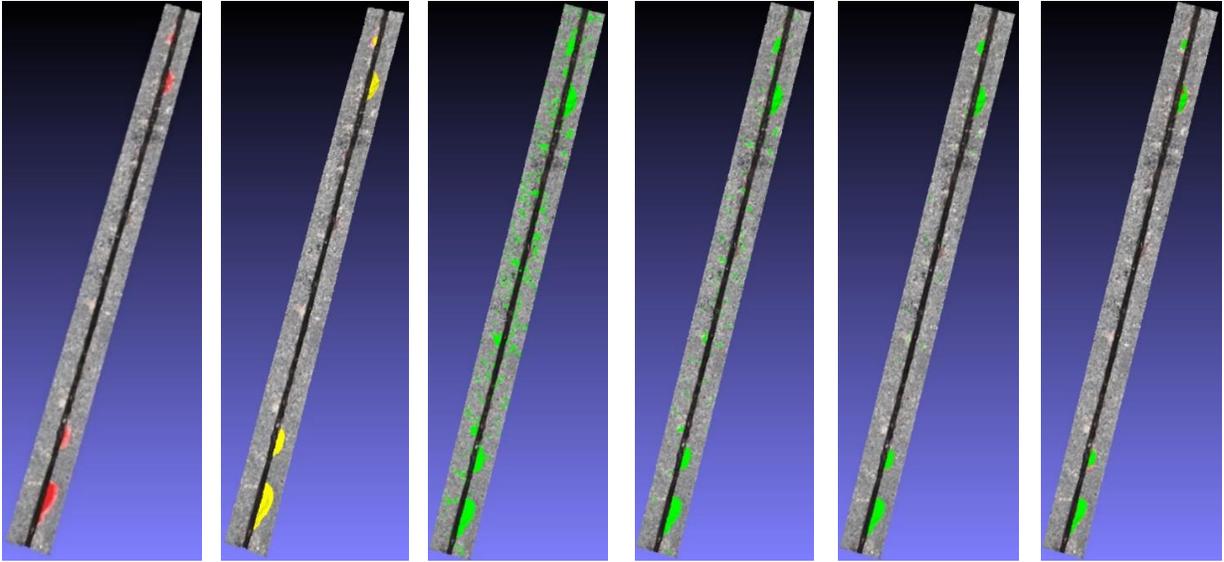

(a)

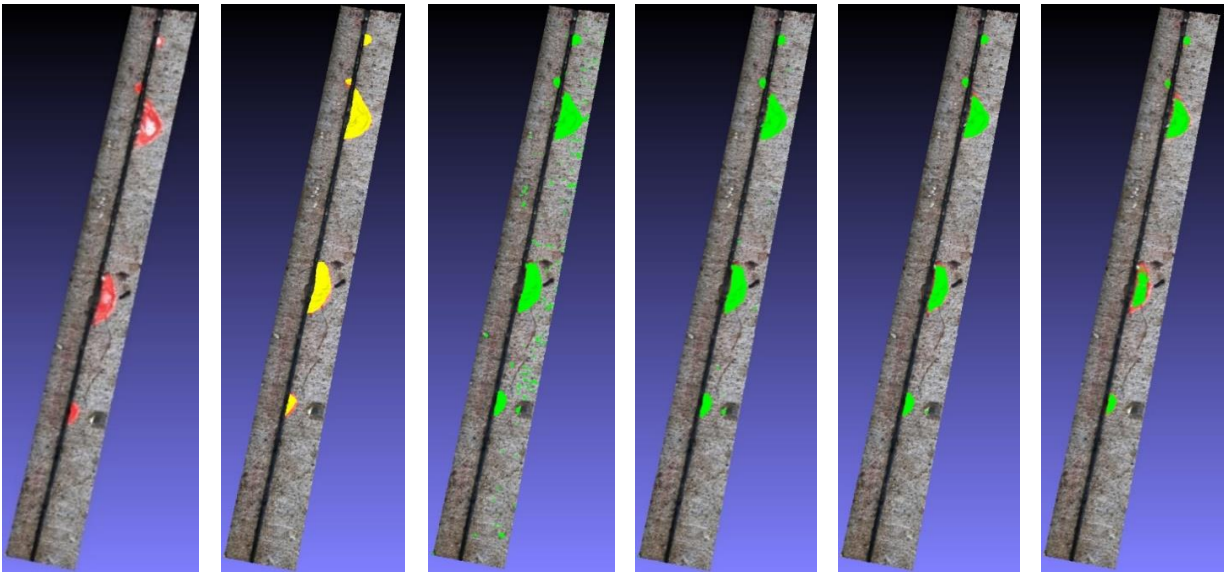

(b)

**Fig. 10** – Influence of color thresholds on joint damage detection for (a) sidewalk and (b) parking lot#1. From left to right, 3D reconstructed image with spall damages in red, the image having ground truth damage (yellow), and next 4 images showing damage detected by color thresholds (green) of 190, 210, 230, and 250, respectively.

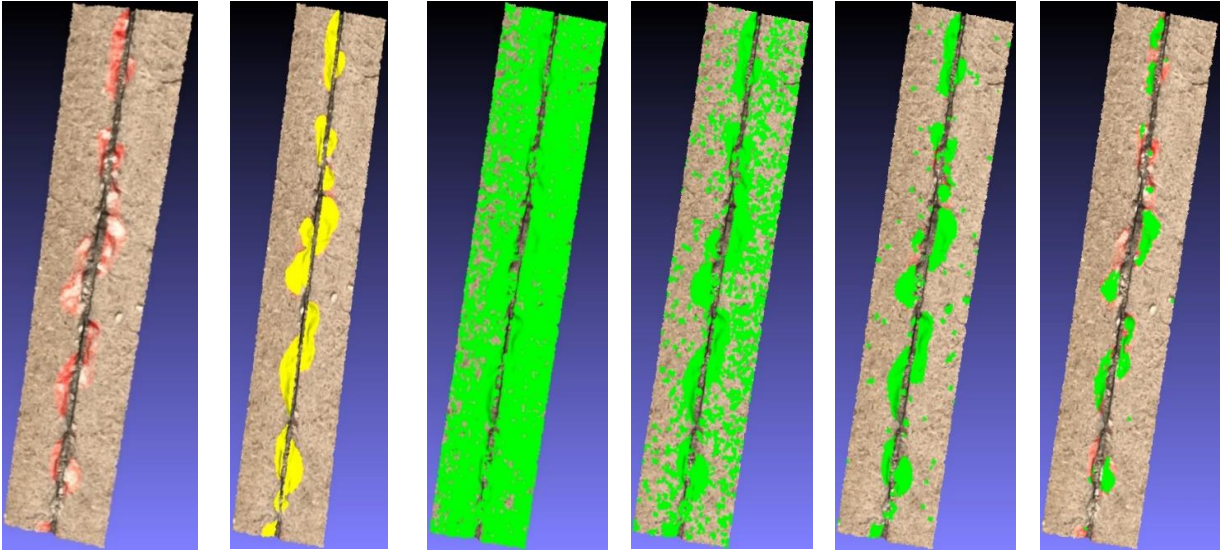

(a)

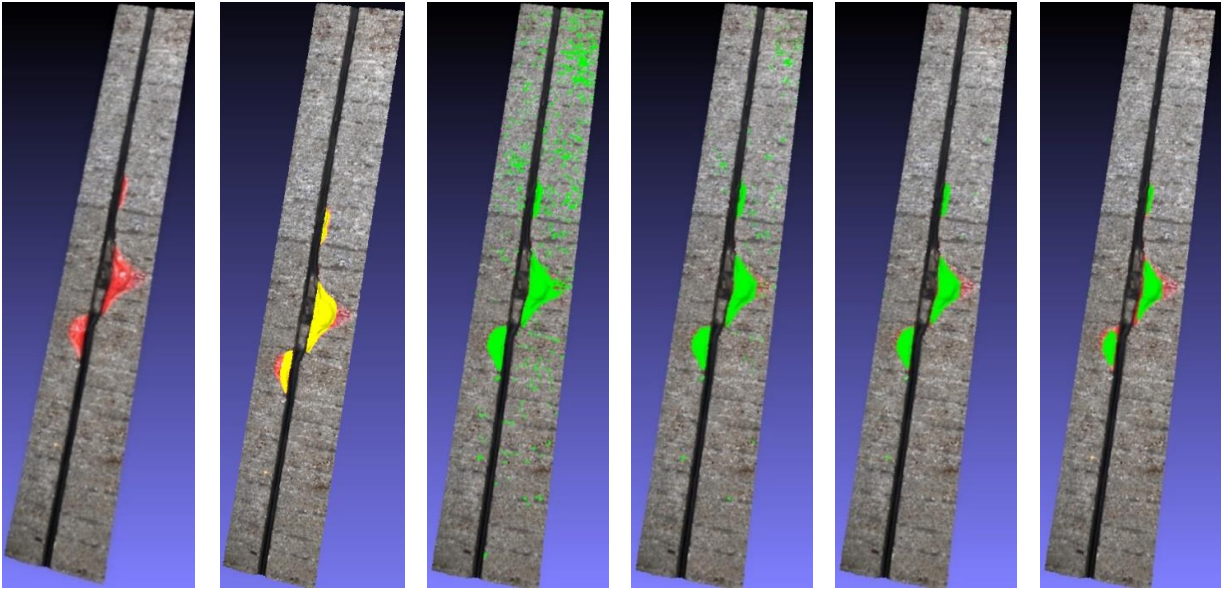

(b)

**Fig. 11** – Influence of color thresholds on joint damage detection for (a) parking lot#2 and (b) JPCP. From left to right, 3D reconstructed image with spall damages in red, the image having ground truth damage (yellow), and next 4 images showing damage detected by color thresholds (green) of 190, 210, 230, and 250, respectively.

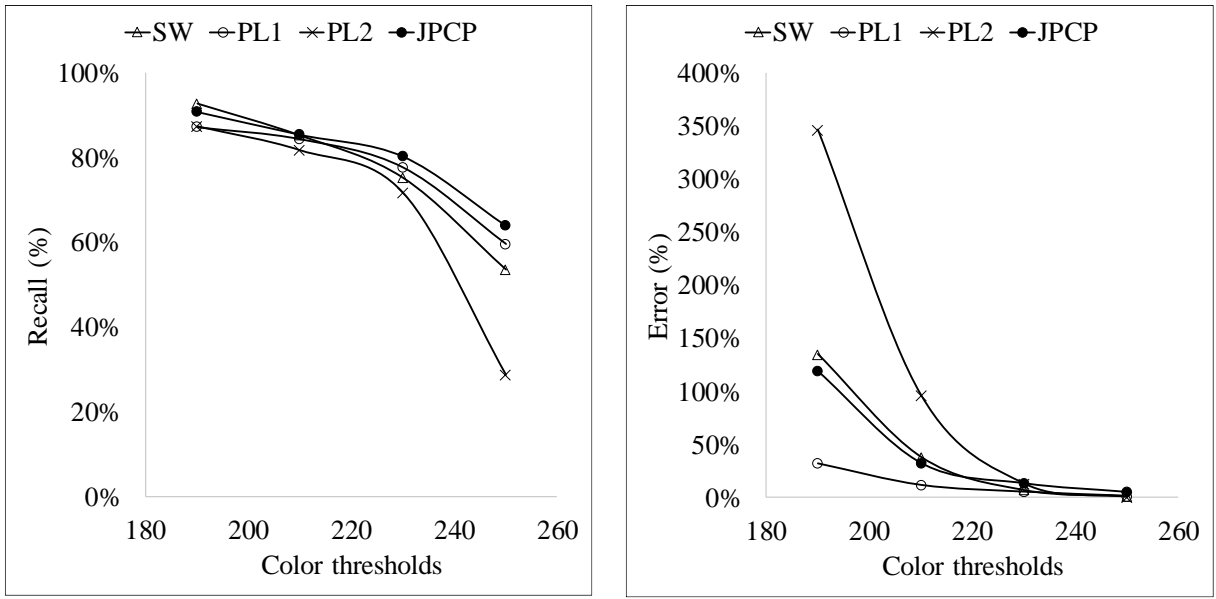

**(a)**                                                                  **(b)**

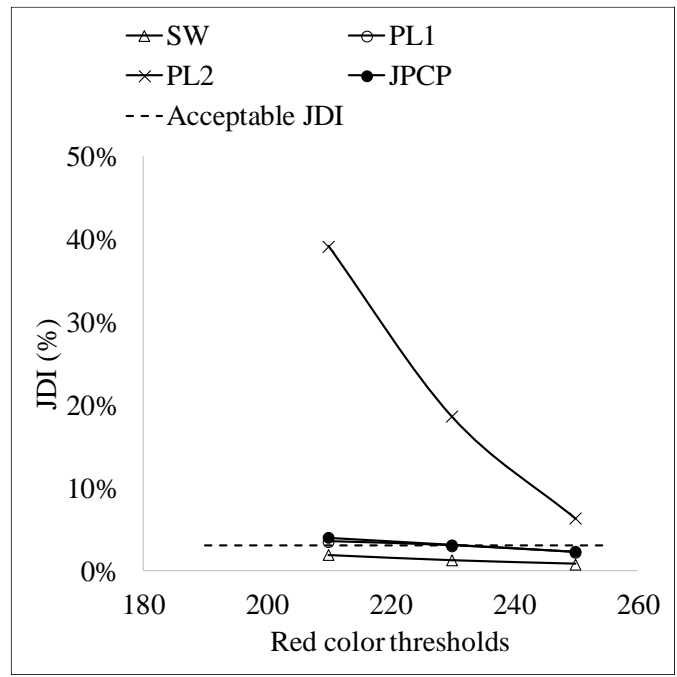

**(c)**

**Fig. 12** – (a) Recall, (b) error, and (c) JDI at different color thresholds for the four contraction joints (Sidewalk-SW, parking lot#1-PL1, parking lot#2-PL2, JPCP). The dashed line represents the acceptable damage index [41].

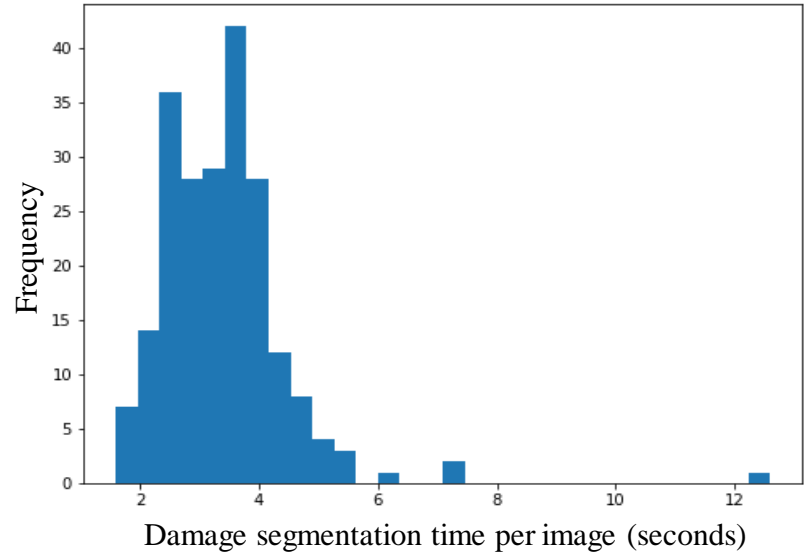

**Fig. 13** – Processing time histogram of 2D damage segmentation.